\documentclass[runningheads]{llncs}

\usepackage{graphicx}
\usepackage{amsmath,amssymb,amsfonts}
\usepackage{algorithmic}
\usepackage{graphicx}
\usepackage{float}
\usepackage{subfig}
\usepackage{subfloat}
\usepackage{textcomp}
\usepackage{xcolor}
\usepackage{caption}
\usepackage{tablefootnote}
\usepackage{url}
\usepackage[export]{adjustbox}
\usepackage{nth}
\usepackage{array, multirow, bigdelim, makecell, booktabs}

\usepackage[flushleft]{threeparttable}

\usepackage[backend=biber,style=numeric,sorting=none]{biblatex}
\begin{document}
\title{Input Layer Binarization with Bit-Plane Encoding}
%
%
\author{Lorenzo Vorabbi\inst{1,2}\orcidID{0000-0002-4634-2044} \and
Davide Maltoni\inst{2}\orcidID{0000-0002-6329-6756} \and
Stefano Santi\inst{1}}
\authorrunning{L. Vorabbi et al.}
%
\institute{Datalogic Labs, Bologna 40012, IT \and
University of Bologna, DISI, Cesena Campus, Cesena 47521, IT
\email{\{lorenzo.vorabbi2,davide.maltoni\}@unibo.it}\\
\email{\{lorenzo.vorabbi,stefano.santi\}@datalogic.com}}
\maketitle              

\begin{abstract}
Binary Neural Networks (BNNs) use 1-bit weights and activations to efficiently execute deep convolutional neural networks on edge devices. Nevertheless, the binarization of the first layer is conventionally excluded, as it leads to a large accuracy loss. The few works addressing the first layer binarization, typically increase the number of input channels to enhance data representation; such data expansion raises the amount of operations needed and it is feasible only on systems with enough computational resources. In this work, we present a new method to binarize the first layer using directly the 8-bit representation of input data; we exploit the standard bit-planes encoding to extract features bit-wise (using depth-wise convolutions); after a re-weighting stage, features are fused again. The resulting model is fully binarized and our first layer binarization approach is model independent. The concept is evaluated on three classification datasets (CIFAR10, SVHN and CIFAR100) for different model architectures (VGG and ResNet) and, the proposed technique outperforms state of the art methods both in accuracy and BMACs reduction.

\keywords{Binary Neural Networks \and Deep Learning}
\end{abstract}
\section{Introduction}
\label{intro}
Deep Neural Networks showed in the last years impressive results, sometimes reaching accuracy better than human level, with applications in a wide variety of domains. These improvements have been achieved by increasing the depth and complexity of the network; such huge models can run smoothly on expensive GPU-based machines but cannot be easily deployed to edge devices (i.e., small mobile or IoT systems), which are typically resource-constrained. Various techniques have been introduced to mitigate this problem, including network quantization \cite{choi2018pact, hubara2016binarized, lin2017towards, rastegari2016xnor, zhou2016dorefa}, network pruning \cite{han2015deep, wen2016learning} and efficient architecture design \cite{howard2017mobilenets, ma2018shufflenet, tan2019efficientnet, tan2021efficientnetv2, hou2021coordinate}.

In 2016, Courbariaux and Bengio \cite{courbariaux2016binarized} first showed the potential of the extreme quantization level that uses only 1-bit to represent both weights and activations. By representing $+1$ with an unset bit and $-1$ with a set bit, the multiplications of weights and activations can be executed with xnor gates, saving hardware resources and greatly reducing power consumption. Such BNN model was able to achieve comparable accuracy results on small datasets like CIFAR10 \cite{krizhevsky2009learning} and SVHN \cite{netzer2011reading} but on wider dataset like Imagenet \cite{ILSVRC15} a relevant accuracy drop was reported. Recent works \cite{rastegari2016xnor, liu2018bi, gu2019projection, xu2019main, qin2020forward, bethge2020meliusnet, bulat2020high, martinez2020training, liu2020reactnet, shi2022repbnn} on BNNs have significantly improved the accuracy on large datasets like Imagenet filling the gap with real-valued networks.

Most of the BNNs do not fully exploit the benefits of 1-bit quantization, since they exclude from binarization the first and last layers that normally work with fixed-point numbers. In general, the number of parameters and the computational effort of the first layer are relatively low compared to intermediate deep convolutional layers employed in VGG \cite{simonyan2014very} or ResNet \cite{he2016deep} models, since input data has typically fewer channels (e.g. color images have three channels). This usually leads to deploy the first layer of BNN models using floating-point or quantizing it using 8-bit; the consequence is that two different type of multipliers (8-bit for first layer, binary for the remaining), with different bit widths, are needed to execute the computations leading to a solution which increases the power consumption (8-bit multiplier requires more power than xnor) and consumes more hardware resources (e.g. an FPGA design) than xnor gates. Conversely, the challenge of binarizing both weights and activations in the input layer is due to the small number of input channels \cite{anderson2017high}. Therefore, almost all the works addressing the binarization of the first layer tried to increase the number of input channels to enrich data representation.

FBNA \cite{anderson2017high} proposes a two-step optimization scheme that consists of binarization and pruning; during binarization phase the number of input channels is increased by a factor $256\times$ and then, during pruning, lowest bits of input data are dropped away. The constraint of FBNA is that the encoded vector must be a power of two. BIL \cite{durichen2018binary} attempts to directly unpack the 8-bit fixed-point input data, called \textit{DBID}, and adding an additional binary pointwise convolutional layer between the unpacked input data and the first layer to increase the number of channels, dubbed as \textit{BIL}. The authors of FracBNN \cite{zhang2021fracbnn} propose to use thermometer encoding to transform a pixel to a thermometer vector (expanding each input channel to 32 binary channels) that then is transformed to the $\left \{ -1,+1 \right \}$ bipolar representation.

\begin{figure}[!t]
	\centering
	\includegraphics[width=0.7\linewidth]{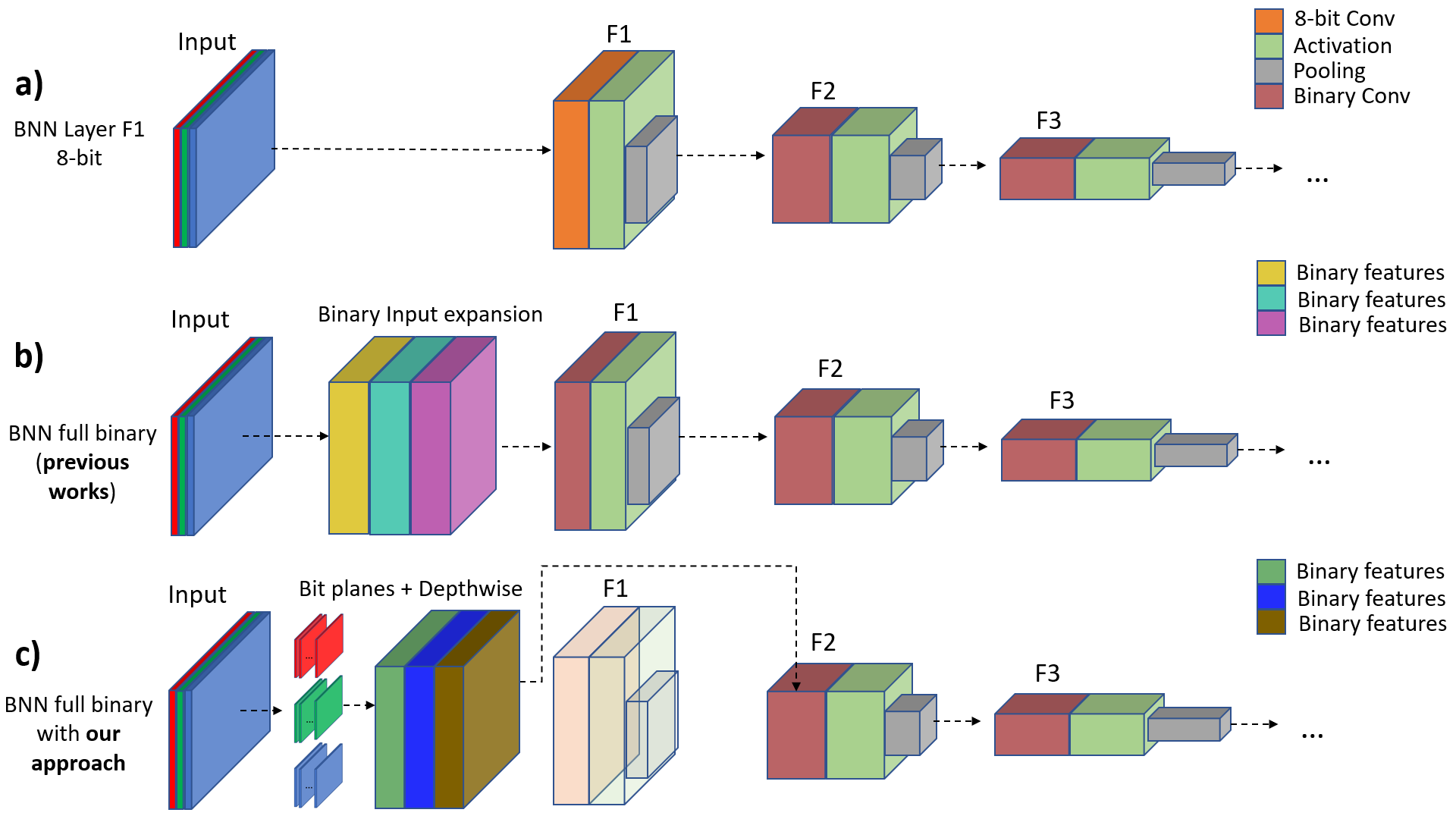}
  	\caption{\textit{(a)} Standard scenario of BNNs where the first convolutional layer is not binarized; weights and inputs are used in 8-bit/floating-point representation. \textit{(b)} Typical approach of the works that binarized the first layer $F_{1}$ incrementing the number of input channels; in this case the input expansion is actually an additional layer. \textit{(c)} Our approach, where depth-wise convolutions are applied to input bit-planes and the resulting maps can replace the $F_{1}$ layer, producing a more compact model.}
  	\label{fig:model_transformation}
\end{figure}

In contrast with previous works where the number of input channels has been increased, our method directly uses the fixed-point representation of a pixel. The results show that the proposed technique is competitive both in term of efficiency and accuracy. Our contributions can be summarized as follows:

\begin{itemize}
\item we propose a general approach to binarize the first layer of a CNN using the native 8-bit fixed-point inputs. We rearrange the 8-bit input data into $8$ bit planes, each bit plane is consumed by a binary depth-wise convolutional layer which gives more importance (using a multiplier, actually a shift operation) to the most significant bit planes. Finally, all feature maps are fused together through an addition operator. The entire process, depicted in Fig. \ref{input_binarization_process}, does not rely on floating-point computation, resulting more suitable to be deployed on ASIC or FPGA systems.

\item we show that the feature maps resulting from our bit-plane manipulations allow to skip the original\footnote{Before the addition of our depth-wise convolutions.} $F_{1}$ first network layer (see Fig. \ref{fig:model_transformation}c) with a minimal accuracy loss, leading to a model which uses less BMACs.

\item we evaluate our concept on three classification datasets (SVHN, CIFAR10 and CIFAR100 \cite{krizhevsky2009learning}) showing that our solution outperforms all previous methods introduced to binarize the input layer.
\end{itemize}

\section{Method}

A common CNN model employed for computer vision problems works with RGB input images; it takes an input volume with three channels ($H \times W \times C$, where $C$ is the number of channels) and extracts the features using convolutional blocks. To increase the receptive field of the network, a sequence of \textit{pooling} operations is used. Each input pixel $p$ is usually a fixed-point integer with $8$ bit precision, namely $p = \sum_{m=0}^{7}x_{m}\cdot 2^{m}$. 

In BNNs, typically the first layer (usually a convolutional one) is not binarized, all the input pixels are processed using 8-bit weights, producing $F_{1}$ output 8-bit feature maps (Fig. \ref{fig:model_transformation}a). 
The previous works in literature that addressed the problem to generate $F_{1}$ binary feature maps, adopted different techniques to increase the number of input channels $C$ ( generating a more sparse representation) in order to use binary weights and inputs for layer $F_{1}$; usually a good tradeoff between accuracy and increment of first layer MACs is to wide the number of channels $C$ by $32 \times$ \cite{durichen2018binary, zhang2021fracbnn}. This process is depicted in Fig. \ref{fig:model_transformation}b, where the increment of input channels leads to a bigger model footprint; in fact, a linear increment of the number of input channels, linearly increases also the kernel parameters of a 2D convolutional layer.

\begin{figure}[!t]
	\centering
	\hfill
	\subfloat[][Bit Planes example]
	{
		\includegraphics[width=0.27\textwidth,valign=M]{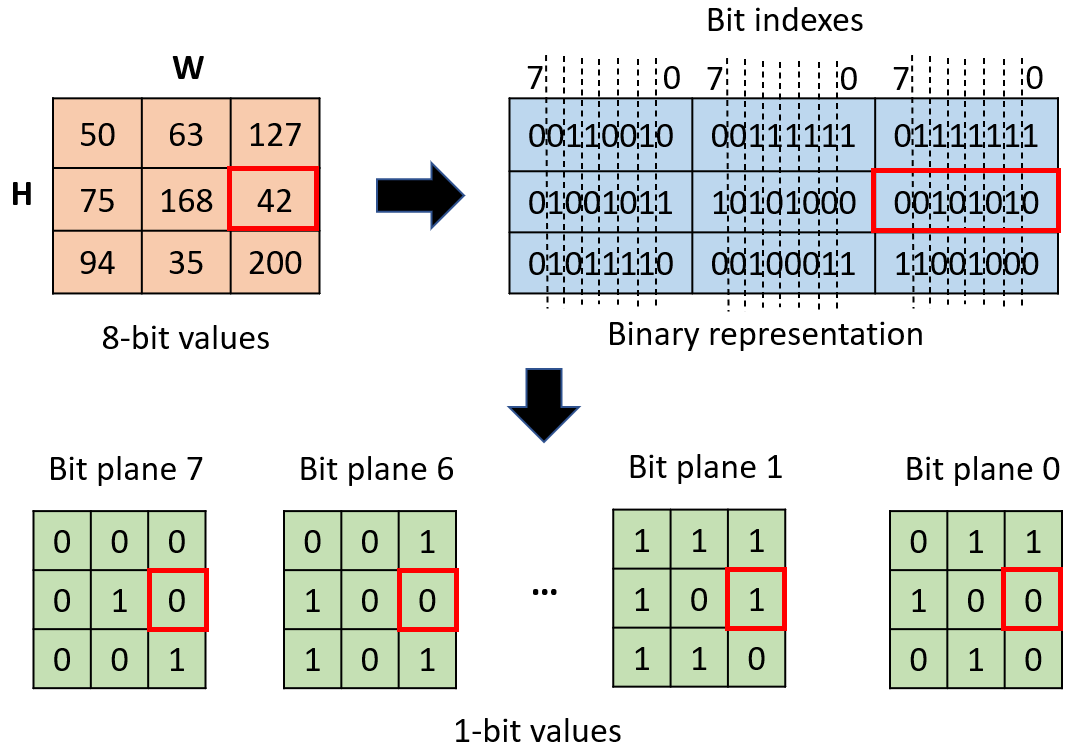}
		\label{fig:bit_planes_example}
	}
	\hfill
	\subfloat[][Bit Planes on CIFAR10, CIFAR100 and SVHN]
	{
		\includegraphics[width=0.66\textwidth,valign=M]{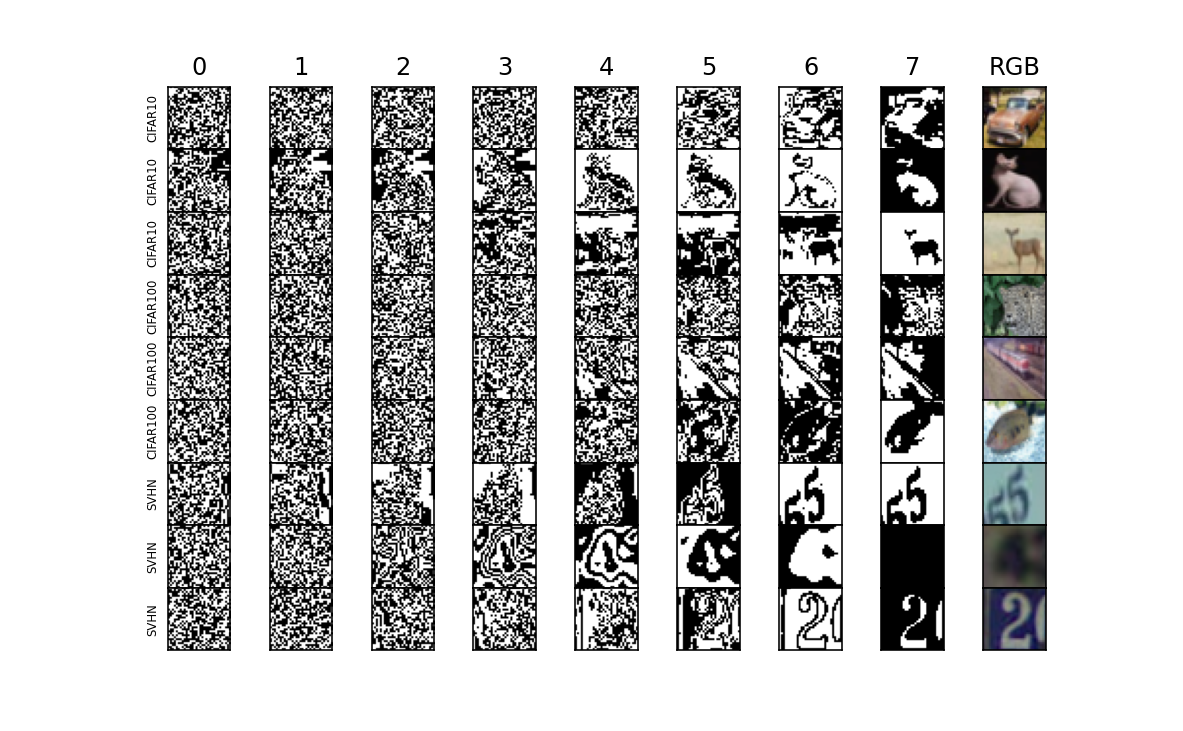}
		\label{fig:bit_planes}
	}
	\hfill
  	\caption{\protect\subref{fig:bit_planes_example} Example of bit plane representation for a $3 \times 3$ 8-bit image. \protect\subref{fig:bit_planes} Image representation in bit planes. Each column refers to a bit index extracted from image; for representation purposes, bit $1$ is converted to $255$ while bit $0$ remains $0$. In this example all bit planes refer to channel G of RGB images.}
  	\label{fig:bit_planes_all}
\end{figure} 

\begin{figure}[!t]
	\centering
	\subfloat[][Bit Rearrangement]
	{
		\includegraphics[width=0.4\textwidth]{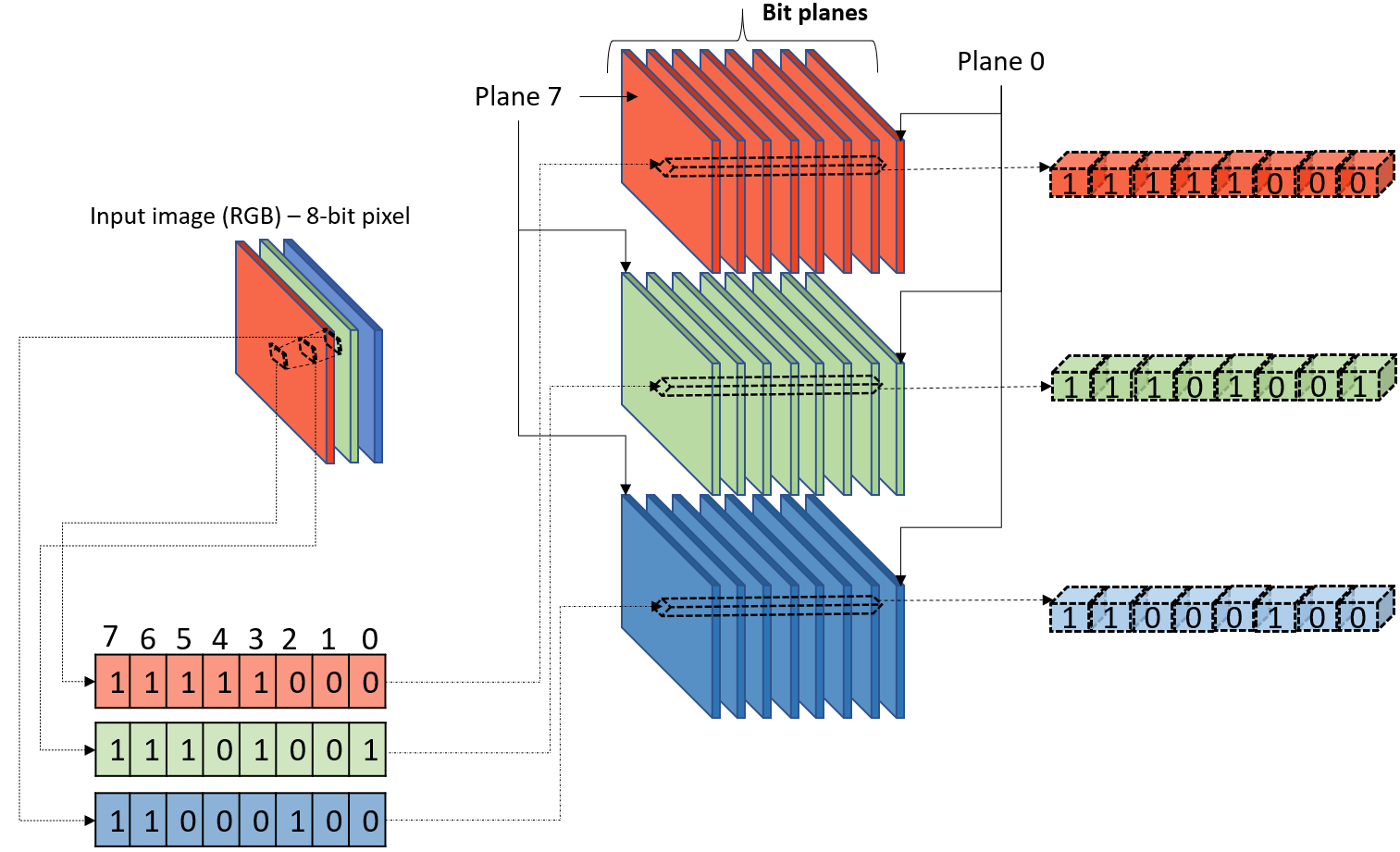}
		\label{fig:bit_rearrangement}
	}
	\hfill
	\subfloat[][Features Extraction]
	{
		\includegraphics[width=0.4\textwidth]{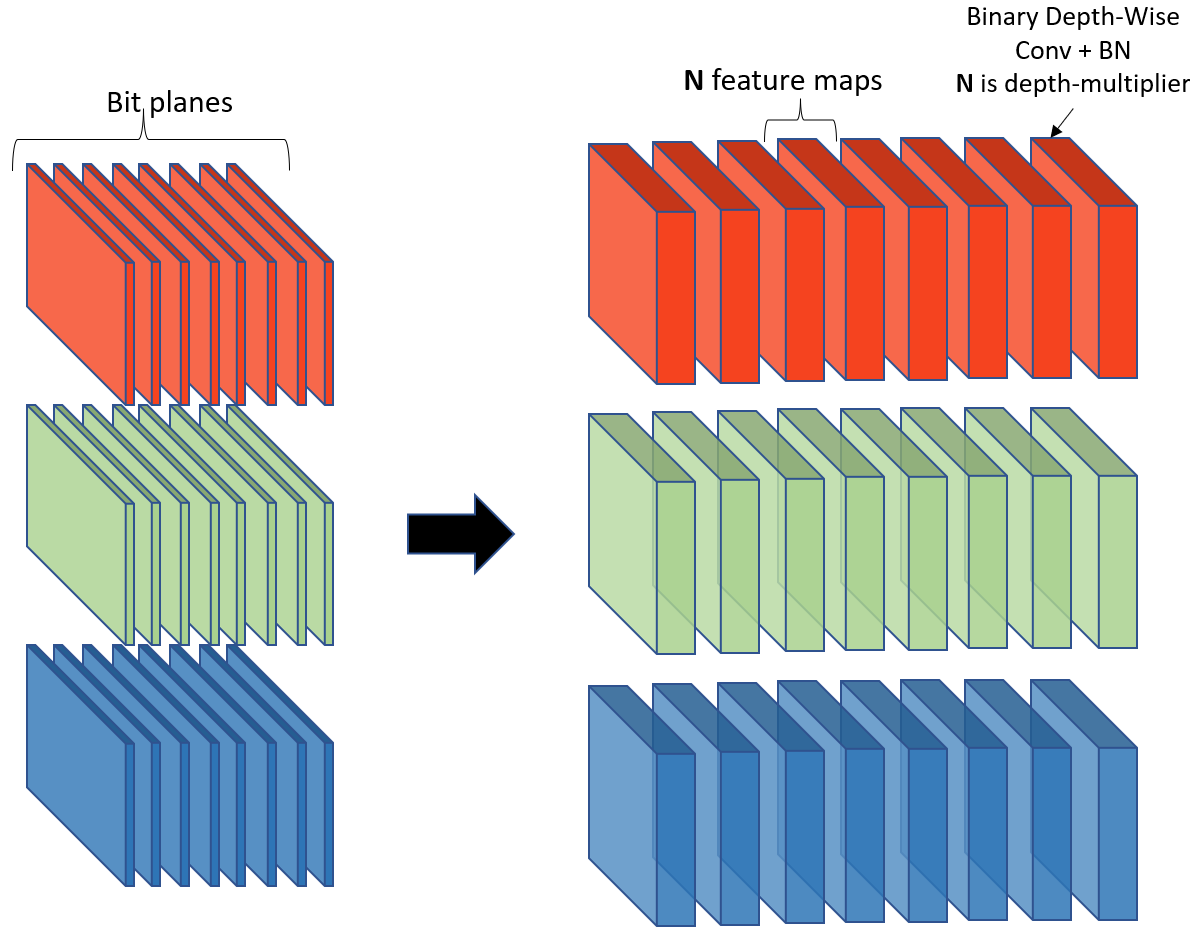}
		\label{fig:feature_extraction}
	}
	\hfill
	\vfill
	\subfloat[][Features Re-Weight]
	{
		\includegraphics[width=0.4\textwidth]{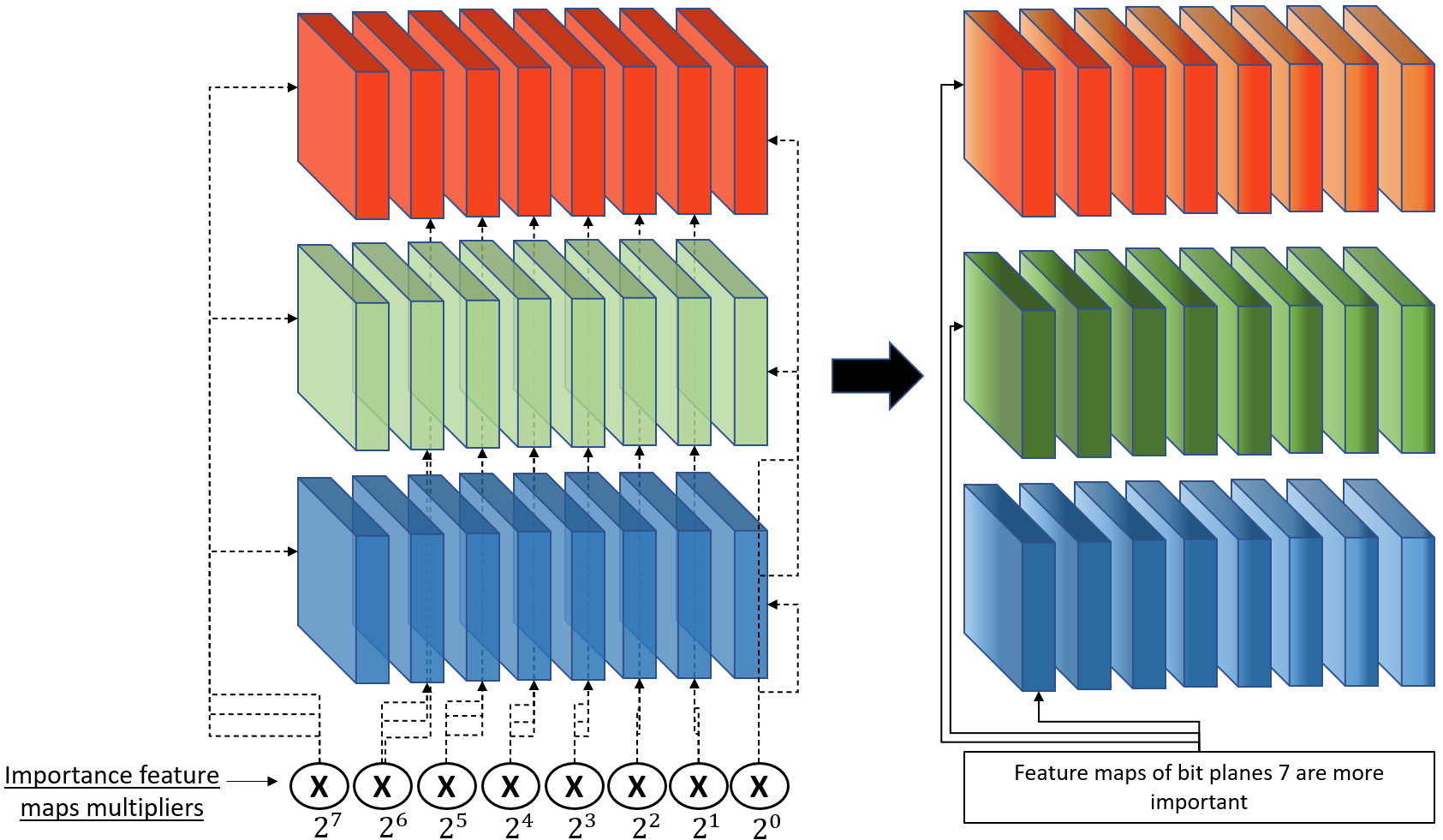}
		\label{fig:features_re_weight}
	}
	\hfill
	\subfloat[][Features Fusion]
	{
		\includegraphics[width=0.4\textwidth]{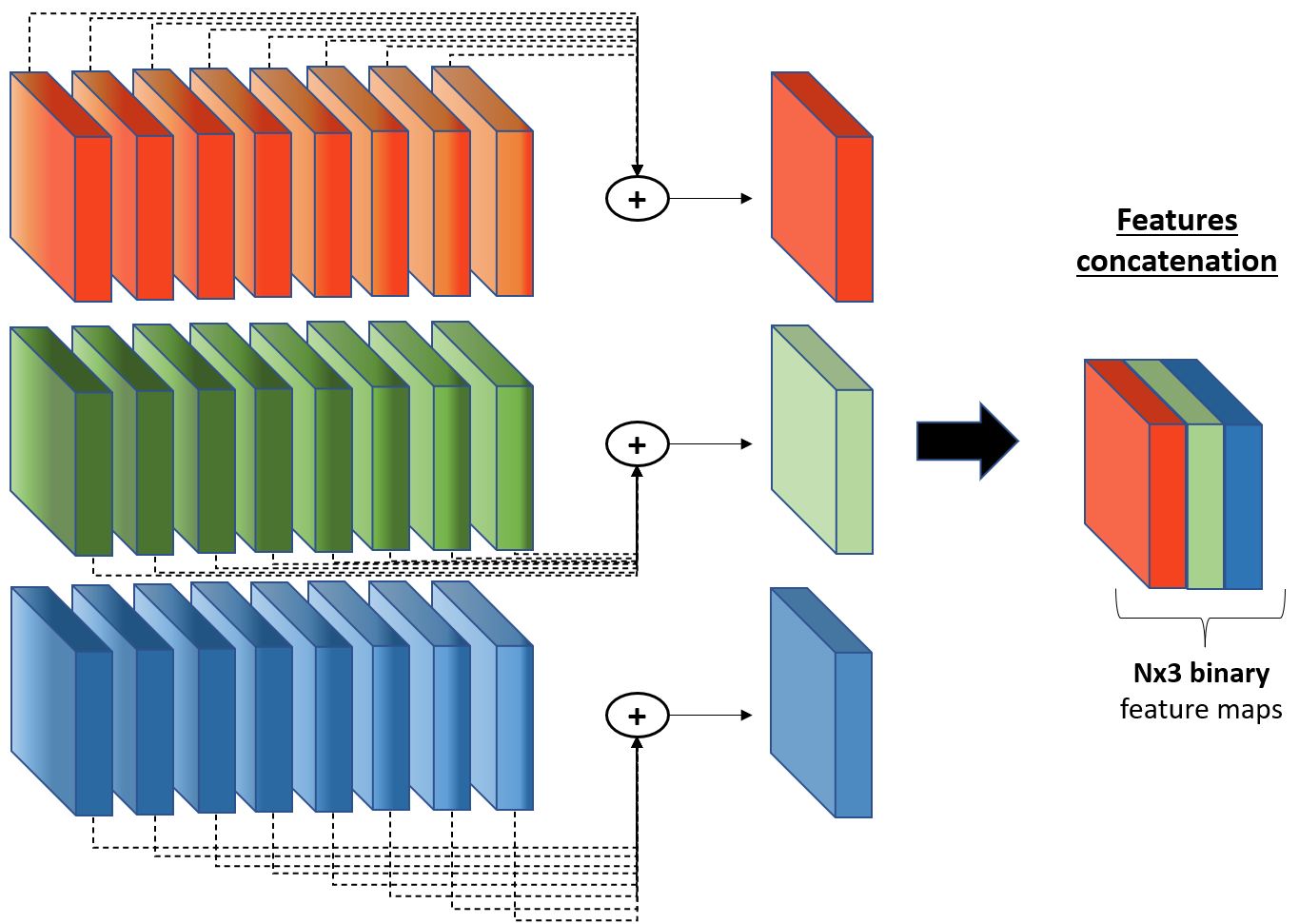}
		\label{fig:features_fusion}
	}
	\hfill
  	\caption{Binarization process of input layer. \protect\subref{fig:bit_rearrangement} shows the rearrangement phase that extracts, for each bit position of the encoded pixel a bit plane. \protect\subref{fig:feature_extraction} shows the binary depth-convolution block applied to each bit plane; the depth multiplier (\textit{N}) is an hyperparameter and it is dataset and model dependent. \protect\subref{fig:features_re_weight} shows how to weight differently feature maps extracted from different bit planes; maps related to most significant bits receive a higher multiplication factor. In \protect\subref{fig:features_fusion}, the feature maps related to the same 8-bit input channel are fused together through an addition.}
    \label{input_binarization_process}
\end{figure}

The intuition behind our approach is that, extracting a different bit plane for each bit position, the semantic spatial information is preserved for most of the high index bits ($4$ to $7$), as shown in Fig. \ref{fig:bit_planes}. Lower bit indexes ($0-3$) contain less correlated spatial information of image pixels and, depending on the dataset, they can be selectively omitted to further reduce the computational effort. The overall diagram of our method is reported in Fig. \ref{input_binarization_process} and it is composed by the following steps:

\begin{enumerate}
\item \textbf{Bit Rearrangement:}
An input image $\mathcal{I}$ ($W, H, C$, where $C$ is the number of channels), having $M$ bits for each pixel (usually $8$), is rearranged into bit planes (as shown in Fig. \ref{fig:bit_rearrangement}); each 8-bit input channel is decomposed into eight 1-bit planes. A bit plane $x$ is a 1-bit map containing only the bit of index $x$ for all pixels (see Fig. \ref{fig:bit_planes_example}). The bit-plane image $bp$ corresponding to channel $c$ can be indicated as $\mathcal{I}\left(c, bp \right)$.

\item \textbf{Feature Extraction:}
Each binary bit-plane is consumed by a binary depthwise convolution layer that generates \textbf{N} feature maps for each bit plane, as reported in Fig. \ref{fig:feature_extraction}. The output of feature extraction ($\mathcal{FE}$) step  can be formulated as:

\begin{equation}
\label{feature_extraction_eq}
\mathcal{FE}\left(c, bp \right) = \gamma\left(c, bp \right) \frac{\left( \mathcal{I}\left(c, bp \right) \ast W\left(c, bp \right) + b\left(c, bp \right) \right) - \mu\left(c, bp \right)}{\sigma\left(c, bp \right)} + \beta\left(c, bp \right)
\end{equation}

where $\ast$ is the convolution operator, $W\left(c, bp \right)$ and $b\left(c, bp \right)$ represent the weights of the depth-wise convolution while $\gamma, \mu, \sigma$ and $\beta$ are the Batch Normalization (BN) \cite{ioffe2015batch} parameters; Eq. \ref{feature_extraction_eq} refers to a single feature map of depth-wise convolution, which is dependent on channel $c$ and bit plane $bp$. In Eq. \ref{feature_extraction_eq} the non-linear activation function can be omitted because binarization of activation and weights already introduce non-linearity. The use of Batch Normalization after each binary layer plays a key role in BNNs because it promotes a smoother optimization process allowing a stable behavior of the gradients. BN layer is usually executed in floating-point precision when mixed with binary layers, but the authors of \cite{vorabbi2023optimizing} proved that it can be executed with 8-bit fixed point without accuracy loss.

\item \textbf{Features Re-Weight:}
Following the intuition based on Fig. \ref{fig:bit_planes}, where high index bit planes preserve the spatial information of the image, this stage re-weights the feature maps based on the bit plane index. Higher bit planes are multiplied by higher scalar values. In order to simplify this stage, the multiplication can be replaced by a shift operation. The \textit{N} feature maps of each bit plane are shifted by the same quantity (namely a power of two).

\item \textbf{Features Fusion:}
The re-weighted feature maps, corresponding to a different 8-bit input channel, are summed to combine the information encoded by different bit indexes and can be expressed as:

\begin{equation}
\label{feature_fusion_eq}
\mathcal{FWF}\left(c, bp \right) = \sum_{i=0}^{M} \mathcal{FE}\left(c, bp \right) \cdot 2^{i}
\end{equation}

In Eq. \ref{feature_fusion_eq} the multiplication by $2^{i}$ represents the re-weight of feature maps that can be implemented with a shift operation.  The sum instead can be implemented accumulating features over 32-bit register; if the subsequent layer extracts \textit{sign} from inputs, then the 32-bit output maps can be reduced to 1-bit saving memory overhead. 
The \textit{N} feature maps corresponding to a different channel are concatenated to create a volume of $N \times 3$ maps that is used to feed the network, Fig. \ref{fig:model_transformation}c. Such volume of $N$ maps can replace the first layer of the CNN with almost no accuracy loss, as showed in Sec. \ref{Resuls_tables} and, thus reducing the complexity of the overall model. $F_{1}$ requires a topology change of the first network layer when its weights and inputs are binarized and the expansion of depth-wise convolutions (Fig. \ref{fig:feature_extraction}) can be set up in order to keep a number of feature maps equivalent to the layer $F_{1}$.

\end{enumerate}

Table \ref{Tab:macs_computation} reports the MACs of different approaches used to binarize input layer of a CNN, reporting the theoretical speedup of the methods; our solution is clearly competitive, in terms of MACs, with respect to the baseline (input not binarized) and other existing approaches.

\begin{table}[]
\centering
\begin{threeparttable}
\begin{tabular}{c|ccc|c|c}
\toprule
 Method & Type & \# MACs & \# weights & $\frac{MACs\: method}{MACs\:Baseline}$\tnote{1} & Latency Speedup\tnote{2} \\
 \hline
 Baseline & 8-bit & $H  W  C  F^{2}  F_{1}$ & $CF^{2}F_{1}$ & $1 \times$ & $1 \times$ \\
 \textit{DBID}\cite{durichen2018binary} & 1-bit & $H  W  C  M  F^{2} F_{1}$ & $CMK + F^{2}F_{1}K$ & $8\times$ & $1.12 \times$ \\
 \textit{BIL}\cite{durichen2018binary} & 1-bit & $H  W  K  \left ( C  M + F^{2}  F_{1} \right )$ & $CKF^{2}F_{1}$ & $10.8 \times$ & $0.81 \times$ \\
 \textit{Thermometer}\cite{zhang2021fracbnn} & 1-bit & $H  W  C  K  F^{2}  F_{1}$ & $CF^{2}N_{1}M$ & $32 \times$ & $0.27 \times$ \\
 \textit{ours $\left ( P=8, N_{1} \right )$} & 1-bit & $H  W  C  P  F^{2}  N_{1}$ & $CPF^{2}N_{1}$ & $2.6 \times$ & $\boldsymbol{3.42} \times$ \\
 \textit{ours $\left ( P=4, N_{1} \right )$} & 1-bit & $H  W  C  P  F^{2}  N_{1}$ & $CMF^{2}N_{1}$ & $1.3 \times$ & $\boldsymbol{6.84 \times}$ \\
 \textit{ours $\left ( P=4, N_{2} \right )$} & 1-bit & $H  W  C  P  F^{2}  N_{2}$ & $CPF^{2}N_{2}$ & $1 \times$ & $\boldsymbol{9 \times}$ \\
 \bottomrule
\end{tabular}
\begin{tablenotes}
	\item[1] A lower ratio means a higher reduction of MACs.
	\item[2] According to \cite{bannink2021larq} (Fig. 2), the worst case speedup of binary convolution compared to 8-bit is $9 \times$.
  \end{tablenotes}
\end{threeparttable}
\caption{Comparison of the first layer MACs required by our method with respect to the state of the art solutions. Input data has a shape $H \times W \times C \left ( 32 \times 32 \times 3 \right )$ and a precision of M bits; in this example the first convolutional layer has $F_{1} = \left (128  \right )$ filters with size $F \times F \left ( 3 \right )$. The expansion channels is $K=32$ for methods \cite{durichen2018binary, zhang2021fracbnn}. The depthwise multiplier of our method can be chosen as $N_{1}=\left \lfloor \frac{F_{1}}{C} \right \rfloor = 42$. We conducted our experiments using also a lower value, $N_{2} = 32$ instead of $N_{1}$ and only $4$ bits of input pixels. $P$ represents the number of bit planes extracted by step \ref{fig:bit_rearrangement}.}
\label{Tab:macs_computation}
\end{table}

\section{Datasets and Implementation Details}

We evaluate our method on three classification datasets: CIFAR10, CIFAR100 and SVHN with different BNN architectures. For each model architecture we tested different state-of-the-art binarization techniques of input layer; input binarization does not modify the other layers of the network, which remain unaltered. For each dataset, we conducted our experiments with the same training procedure (same number of epochs, optimizer, learning rate scheduling, loss function) for all topologies without adding distillation losses or special regularization to the overall loss function. The binarization of weights and activations always happens at training time using an approximation of the gradient (STE \cite{bengio2013estimating} or derived solution that are model dependent) for \textit{sign} function. The augmentation procedure for all datasets is performed with floating-point arithmetic but, before feeding data to the network, input image is quantized using 8-bit fixed precision.
\\
We adopted the following datasets:

\begin{description}
\item[CIFAR10 and CIFAR100] \hspace{0.05\textwidth} The RGB images are scaled to the interval $\left [-1.0\: ;+1.0  \right ]$ and the following data augmentation was used: zero padding of $4$ pixels for each size, a random $32 \times 32$ crop and a random horizontal flip. No augmentation is used at test time. The models have been trained for 140 epochs.
\item[SVHN] \hspace{0.05\textwidth} The RGB input images are scaled to the interval $\left [-1.0\: ;+1.0  \right ]$ and the following data augmentation procedure is used: random rotation ($\pm8$ degrees), zoom ($\left[0.95, 1.05 \right]$), random shift ($\left[0;10\right]$) and random shear ($\left[0;0.15\right]$). The models have been trained for 70 epochs.
\end{description}

We evaluated the following networks:

\begin{description}
\item[VGG-Small\cite{zhang2018lq}] \hspace{0.05\textwidth} Network structure is the following: $2 \times \left ( 128-C3 \right ) + MP2 + 2 \times \left(256-C3 \right) + MP2 + 2 \times \left(512-C3 \right) + MP2 + FC1024 + FC1024 + Softmax$\footnote{$m \times \left ( n-CK \right )$ stands for $m$ consecutive convolutional layers, each one with $n$ output channels and $K$ kernel size. $MP2$ is the max pooling layer with subsample $2$ while $FCx$ is a fully-connected layer having $x$ neurons. $Softmax$ represents the last dense classification layer using softmax as activation.}.
The VGG-Small model adopted uses the straight-through-estimator (STE) to approximate the gradient on non-differentiable layers\cite{hubara2016binarized, bengio2013estimating}.
\item[VGG-11\cite{xu2019main}] \hspace{0.05\textwidth} Network structure is the following: $64-C3 + MP2 + 128-C3 + MP2 + 2 \times \left(256-C3 \right) + MP2 + 2 \times \left(512-C3 \right) + MP2 + 2 \times \left(512-C3 \right) + MP2 + Softmax$. Even VGG-11 uses the STE estimator for binarization operation during back-propagation.
\item[BiRealNet\cite{liu2018bi}] \hspace{0.05\textwidth} It is a modified version of classical ResNet that proposes to preserve the real activations before the sign function to increase the representational capability of the 1-bit CNN, through a simple shortcut. Bi-RealNet adopts a tight approximation to the derivative of the non-differentiable sign function with respect to activation and a magnitude-aware gradient to update weight parameters. We used two instances of the network, an \textit{18-layer} and a \textit{34-layer} Bi-Real net \footnote{Refer to the following \url{https://github.com/liuzechun/Bi-Real-net} repository for all the details.}.
\item[ReactNet\cite{liu2020reactnet}] \hspace{0.05\textwidth} To further compress compact networks, this model constructs a baseline based on MobileNetV1 \cite{howard2017mobilenets} and add shortcut to bypass every 1-bit convolutional layer that has the same number of input and output channels. The $3 \times 3$ depth-wise and the $1 \times 1$ point-wise convolutional blocks of MobileNet are replaced by the $3 \times 3$ and $1 \times 1$ vanilla convolutions in parallel with shortcuts in React Net\footnote{Refer to the following \url{https://github.com/liuzechun/ReActNet} repository for all the details.}. As for Bi-Real Net, we tested two different versions of React Net: a \textit{18-layer} and a \textit{34-layer}.
\end{description}

\section{Results and Conclusions}
\label{Resuls_tables}

\begin{table}[h]
\centering
\noindent\makebox[0.9\textwidth]{
\begin{tabular}{c|c c c c c cl}
\toprule
  & VGG-Small & VGG-11 & BiRealNet-18 & BiRealNet-34 & ReactNet-18 & ReactNet-34 \\
 \midrule

\textit{DBID}\cite{durichen2018binary}$\left ( P=8 \right )$ & $84.5$ & $77.6$ & $81.8$ & $85.0$ & $86.0$ & $86.7$  & \hspace{-1em}\rdelim\}{5}{*}[$\nth{1}$ scenario] \\
\textit{BIL}\cite{durichen2018binary}$\left(P=8\right)$ & $82.0$ & $78.9$ & $81.3$ & $82.2$ & $84.0$ & $83.5$ \\
\textit{Therm}\cite{zhang2021fracbnn}$\left(K=32\right)$ & $84.2$ & $80.1$ & $85.2$ & $85.3$ & $86.6$ & $86.6$ \\
\textbf{ours}$\left ( P=8, N_{1} \right )$ & $\boldsymbol{85.9}$ & $79.1$ & $\boldsymbol{87.7}$ & $\boldsymbol{88.5}$ & $\boldsymbol{89.9}$ & $\boldsymbol{90.2}$ \\
\textit{baseline} & $89.2$ & $84.7$ & $89.1$ & $89.3$ & $90.6$ & $90.6$ \\
\hline
\\
\hline
\textit{DBID}$\left ( P=4\right )$ & $83.6$ & $76.8$ & $74.9$ & $83.7$ & $83.7$ & $85.3$ & \hspace{-1em}\rdelim\}{5}{*}[$\nth{2}$ scenario] \\
\textit{BIL}$\left ( P=4 \right )$ & $80.9$ & $82.4$ & $80.8$ & $82.3$ & $82.7$ & $83.4$ \\
\textit{Therm}$\left ( K=16\right )$ & $83.8$ & $79.6$ & $84.7$ & $85.9$ & $86.5$ & $86.8$ \\
\textbf{ours}$\left ( P=4, N_{2} \right )$ & $\boldsymbol{85.0}$ & $78.3$ & $\boldsymbol{86.9}$ & $\boldsymbol{87.7}$ & $\boldsymbol{88.5}$ & $\boldsymbol{89.0}$  \\
\textit{baseline} & $88.3$ & $83.7$ & $87.4$ & $88.3$ & $88.8$ & $89.1$ \\
 \bottomrule
\end{tabular}
}
\caption{Top1 accuracy (\%) results of test set on CIFAR10. In first part we report the result of first test scenario (standard conditions); in second half, the results achieved in the second scenario (reducing the MACs of binarization of input layer).}
\label{Tab:CIFAR10_results}
\end{table}

\begin{table}[h]
\centering
\noindent\makebox[0.9\textwidth]{
\begin{tabular}{c|c c c c c cl}
\toprule
  & VGG-Small & VGG-11 & BiRealNet-18 & BiRealNet-34 & ReactNet-18 & ReactNet-34 \\
 \midrule
\textit{DBID}\cite{durichen2018binary}$\left ( P=8 \right )$ & $94.5$ & $92.2$ & $94.3$ & $95.1$ & $94.9$ & $95.1$ & \hspace{-1em}\rdelim\}{5}{*}[$\nth{1}$ scenario]  \\
\textit{BIL}\cite{durichen2018binary}$\left(P=8\right)$ & $93.5$ & $92.1$ & $94.3$ & $93.4$ & $94.1$ & $94.7$ \\
\textit{Therm}\cite{zhang2021fracbnn}$\left(K=32\right)$ & $89.7$ & $88.9$ & $89.2$ & $89.8$ & $89.8$ & $90.2$ \\
\textbf{ours}$\left ( P=8, N_{1} \right )$ & $\boldsymbol{94.8}$ & $\boldsymbol{93.4}$ & $94.3$ & $95.0$ & $\boldsymbol{95.1}$ & $\boldsymbol{95.7}$  \\
\textit{baseline} & $95.7$ & $95.5$ & $94.3$ & $95.1$ & $95.5$ & $95.9$ \\
\hline
\\
\hline
\textit{DBID}$\left ( P=4\right )$ & $94.3$ & $92.1$ & $94.3$ & $94.7$ & $94.8$ & $95.0$ & \hspace{-1em}\rdelim\}{5}{*}[$\nth{2}$ scenario] \\
\textit{BIL}$\left ( P=4 \right )$ & $93.4$ & $92.1$ & $94.4$ & $93.5$ & $93.8$ & $94.5$ \\
\textit{Therm}$\left ( K=16\right )$ & $89.5$ & $88.6$ & $89.8$ & $89.7$ & $89.8$ & $90.1$ \\
\textbf{ours}$\left ( P=4, N_{2} \right )$ & $\boldsymbol{94.8}$ & $\boldsymbol{93.3}$ & $94.3$ & $\boldsymbol{95.0}$ & $\boldsymbol{95.1}$ & $\boldsymbol{95.7}$  \\
\textit{baseline} & $95.6$ & $95.0$ & $94.4$ & $95.1$ & $95.5$ & $96.0$  \\
 \bottomrule
\end{tabular}
}
\caption{Top1 accuracy (\%) results of test set on SVHN.}
\label{Tab:SVHN_results}
\end{table}

\begin{table}[h]
\centering
\noindent\makebox[0.9\textwidth]{
\begin{tabular}{c|c c c c c cl}
\toprule
  & VGG-Small & VGG-11 & BiRealNet-18 & BiRealNet-34 & ReactNet-18 & ReactNet-34 \\
 \midrule
\textit{DBID}\cite{durichen2018binary}$\left ( P=8 \right )$ & $53.6$ & $43.1$ & $51.8$ & $58.5$ & $56.3$ & $58.0$ & \hspace{-1em}\rdelim\}{5}{*}[$\nth{1}$ scenario]  \\
\textit{BIL}\cite{durichen2018binary}$\left(P=8\right)$ & $50.0$ & $42.9$ & $52.7$ & $56.0$ & $55.4$ & $55.5$ \\
\textit{Therm}\cite{zhang2021fracbnn}$\left(K=32\right)$ & $53.0$ & $43.5$ & $57.2$ & $57.1$ & $57.4$ & $57.9$ \\
\textbf{ours}$\left ( P=8, N_{1} \right )$ & $\boldsymbol{56.5}$ & $\boldsymbol{46.0}$ & $\boldsymbol{58.7}$ & $\boldsymbol{60.6}$ & $\boldsymbol{61.7}$ & $\boldsymbol{62.9}$ \\
\textit{baseline} & $60.6$ & $52.3$ & $63.4$ & $65.0$ & $64.9$ & $65.3$ \\
\hline
\\
\hline
\textit{DBID}$\left ( P=4\right )$ & $52.3$ & $41.8$ & $50.5$ & $56.5$ & $55.2$ & $56.7$ & \hspace{-1em}\rdelim\}{5}{*}[$\nth{2}$ scenario] \\
\textit{BIL}$\left ( P=4 \right )$ & $49.5$ & $42.0$ & $52.1$ & $54.5$ & $52.1$ & $53.6$ \\
\textit{Therm}$\left ( K=16\right )$ & $52.1$ & $42.6$ & $56.7$ & $54.5$ & $56.8$ & $58.6$ \\
\textbf{ours}$\left ( P=4, N_{2} \right )$ & $\boldsymbol{54.8}$ & $\boldsymbol{44.5}$ & $\boldsymbol{57.7}$ & $\boldsymbol{59.6}$ & $\boldsymbol{60.2}$ & $\boldsymbol{62.0}$  \\
\textit{baseline} & $60.3$ & $50.3$ & $60.0$ & $61.7$ & $62.0$ & $63.4$ \\
 \bottomrule
\end{tabular}
}
\caption{Top1 accuracy (\%) results of test set on CIFAR100.}
\label{Tab:CIFAR100_results}
\end{table}

The validation of our solution has been accomplished through two different test scenarios; in the first one, we compared the accuracy (measured on test set) of our binarization method w.r.t. the state-of-the-arts input layer binarization approaches, keeping unaltered the structure of the network except for the input data binarization layer (first half of Tables \ref{Tab:CIFAR10_results}, \ref{Tab:SVHN_results} and \ref{Tab:CIFAR100_results}). In this first scenario all the 8-bits planes are exploited, layer F1 (Fig. \ref{fig:model_transformation}) is executed and our proposed solution is able to reach a better accuracy compared to other input binarization methods, closing the accuracy gap with the baseline.

In the second scenario, to further reduce the MACs of our solution, we propose an optimization of our method that uses only the $4$ most significant bits and reduces the depth-wise multiplier from $N_{1}$ to $N_{2}$ (the reduction to $4$ bits is based on Fig. \ref{fig:bit_planes}, that shows how the bit planes corresponding to less significant bits convey less information). In the second half of tables \ref{Tab:CIFAR10_results}, \ref{Tab:SVHN_results} and \ref{Tab:CIFAR100_results}, we report the results of the optimized version compared with other solutions properly modified in order to compute an equivalent number of channels \footnote{ For \textit{DBID}, \textit{thermometer} and \textit{baseline} methods, we reduced to $32$ the number of output channels of layer $F1$; for \textit{BIL} and \textit{ours}, we skipped the layer $F1$ because the convolution operation is already exploited within the input layer binarization process. For \textit{DBID}, \textit{BIL} and \textit{ours} we used only the $4$ most significant bits of input data. For \textit{thermometer} we applied also a reduced expansion factor of $K=16$.}. As reported, our solution is able to preserve the baseline accuracy using less input bits while the other methods get a consistent accuracy drop when reducing input bits and binary channels. 
\\
Differently from other works, our solution re-weights the feature extracted by bit-planes giving more importance to the features corresponding to the most significant bit-planes; this stage contributes to scale down the footprint of our binarization approach simplifying the deployment on resource constrained devices (low-power embedded CPUs).
Furthermore, the accuracy of our method is higher than \textit{thermometer encoding}\cite{zhang2021fracbnn}, which preserves the feature similarity after binarizing the input layer, as pointed out  by Anderson et al. \cite{anderson2017high}. 

In conclusion, this paper introduced a novel input layer binarization method that reaches higher accuracy when compared to state-of-the-art solutions reducing the gap to the baseline on average by $2.2$ percentage points. Our solution was able to preserve model accuracy when only $4$ bits of input pixels are used in the input binarization layer, proving to be more resource-constrained device friendly than existing ones.
In the future, we intend to further investigate the latency speedup of our method on real hardware devices like Raspberry Pi Model 3B/4B exploiting the computation capabilities of NEON ARM\footnote{\url{https://www.arm.com/technologies/neon}} SIMD engine.

\printbibliography
\end{document}